\documentclass[runningheads]{llncs}
\usepackage{graphicx}
\usepackage{amsmath,amssymb} 
\usepackage{color}
\usepackage{epstopdf}
\usepackage{subfigure}
\usepackage{multirow}
\usepackage{colortbl}
\usepackage{booktabs}
\usepackage{array}
\usepackage{floatrow}
\usepackage[ruled,vlined]{algorithm2e}

\newfloatcommand{capbtabbox}{table}[][\FBwidth]
\begin{document}
\title{Deep Metric Learning with Hierarchical \\ Triplet Loss}

%
\author{Weifeng Ge$^{1,2,3}$ \and
Weilin Huang$^{1,2}$\thanks{Weilin Huang is the corresponding author (e-mail:whuang@malong.com).} \and
Dengke Dong$^{1,2}$ \and Matthew R. Scott$^{1,2}$}
\authorrunning{W. Ge \and W. Huang \and
D. Dong \and M. R. Scott}
%

\institute{$^1$Malong Technologies, Shenzhen, China \\
$^2$Shenzhen Malong Artificial Intelligence Research Center, Shenzhen, China\\
$^3$The University of Hong Kong\\
\email{\{terrencege,whuang,dongdk,mscott\}@malong.com}}
%
\maketitle              
	\begin{abstract}
We present a novel hierarchical triplet loss (HTL) capable of automatically collecting informative training samples (triplets) via a defined hierarchical tree that encodes global context information. This allows us to cope with the main limitation of random sampling in training a conventional triplet loss, which is a central issue for deep metric learning. Our main contributions are two-fold. (i) we construct a hierarchical class-level tree where neighboring classes are merged recursively. The hierarchical structure naturally captures the intrinsic data distribution over the whole dataset. (ii) we formulate the problem of triplet collection by introducing a new violate margin, which is computed dynamically based on the designed hierarchical tree. This allows it to automatically select meaningful hard samples with the guide of global context. It encourages the model to learn more discriminative features from visual similar classes, leading to faster convergence and better performance.
Our method is evaluated on the tasks of image retrieval and face recognition, where it outperforms the standard triplet loss substantially by 1\%–18\%. It achieves new state-of-the-art performance on a number of benchmarks, with much fewer learning iterations.


\keywords{Deep Metric Learning \and  Image Retrieval \and Triplet Loss \and Anchor-Neighbor Sampling}
\end{abstract}

	\section{Introduction}
Distance metric learning or similarity learning is the task of learning a distance function over images in visual understanding tasks. It has been an active research topic in computer vision community. Given a similarity function, images with similar content are projected onto neighboring locations on a manifold, and images with different semantic context are mapped apart from each other.
With the boom of deep neural networks (DNN), metric learning has been turned from learning distance functions  to learning deep feature embeddings that better fits a simple distance function, such as Euclidean distance or cosine distance. Metric learning with DNNs is referred as deep metric learning, which has recently achieved great success in numerous visual understanding tasks, including images or object retrieval \cite{song2016deep,ustinova2016learning,wohlhart2015learning}, single-shot object classification \cite{ustinova2016learning,wohlhart2015learning,waltner2016bacon}, keypoint descriptor learning \cite{kumar2016learning,simo2015discriminative}, face verification \cite{schroff2015facenet,parkhi2015deep}, person re-identification \cite{ustinova2016learning,shi2016embedding}, object tracking \cite{tao2016siamese} and etc.

Recently, there is a number of widely-used loss functions developed for deep metric learning, such as contrastive loss \cite{sun2014deep,hadsell2006dimensionality}, triplet loss \cite{schroff2015facenet} and quadruplet loss \cite{Chen_2017_CVPR}. These loss functions are calculated on correlated samples, with a common goal of encouraging samples from the same class to be closer, and pushing samples of different classes apart from each other, in a projected feature space. The correlated samples are grouped into contrastive pairs, triplets or quadruplets, which form the training samples for these loss functions on deep metric learning.
Unlike softmax loss used for image classification, where the gradient is computed on each individual sample, the gradient of a deep metric learning loss often depends heavily on multiple correlated  samples.
Furthermore, the number of training samples will be increased exponentially when the training pairs, triplets or quadruplets are grouped. This generates a vast number of training samples which are highly redundant and less informative. Training that uses random sampling from them can be overwhelmed by redundant samples, leading to slow convergence and inferior performance.


Deep neural networks are commonly trained using online stochastic gradient descent algorithms \cite{orr2003neural}, where the gradients for optimizing network parameters are computed \emph{locally} with mini-batches, due to the limitation of computational power and memory storage. It is difficult or impossible to put all training samples into a single mini-batch, and  the networks can only focus on local data distribution within a mini-batch, making it difficult to consider global data distribution over the whole training set. This often leads to local optima and slow convergence. This common challenge will be amplified substantially in deep metric learning, due to the enlarged sample spaces where the redundancy could become more significant. Therefore, collecting and creating meaningful training samples (e.g., in pairs, triplets or quadruplets) has been a central issue for deep metric learning, and an efficient sampling strategy is of critical importance to this task. This is also indicated in recent literature \cite{schroff2015facenet,Wu_2017_ICCV,parkhi2015deep,amos2016openface}.

Our goal of this paper is to address the sampling issue of conventional triplet loss \cite{schroff2015facenet}. In this work, we propose a novel hierarchical triplet loss (HTL) able to automatically collect informative training triplets via an adaptively-learned hierarchical class structure that encodes global context in an elegant manner. Specifically, we explore the underline data distribution on a manifold sphere, and then use this manifold structure to guide triplet sample generation. Our intuition of generating meaningful samples is to encourage the training samples within a mini-batch to have similar visual appearance but with different semantic content (e.g., from different categories). This allows our model to learn more discriminative features by identifying subtle distinction between the close visual concepts. Our main contribution are described as follows.

--- We propose a novel hierarchical triplet loss that allows the model to collect informative training samples with the guide of a global class-level hierarchical tree. This alleviates main limitation of random sampling in training of deep metric learning, and encourages the model to learn more discriminative  features from visual similar classes.

--- We formulate the problem of triplet collection by introducing a new violate margin, which is computed dynamically over the constructed hierarchical tree. The new violate margin allows us to search informative samples, which are hard to distinguish between visual similar classes, and will be merged into a new class in next level of the hierarchy. The violate margin is automatically updated, with the goal of identifying a margin that generates gradients for violate triplets, naturally making the collected samples  more informative.

--- The proposed HTL is easily implemented, and can be readily integrated into the standard triplet loss or other deep metric learning approaches, such as contrastive loss, quadruplet loss,  recent HDC \cite{Yuan_2017_ICCV} and BIER \cite{Opitz_2017_ICCV}. It significantly outperforms the standard triplet loss on the tasks of image retrieval and face recognition, and obtains new state-of-art results on a number of benchmarks.

	\section{Related work}
\noindent\textbf{Deep Metric Learning.} Deep metric learning maps an image into a feature vector in a manifold space via deep neural networks. In this manifold space, the Euclidean distance (or the cosine distance) can be directly used as the distance metric between two points. The contribution of many deep metric learning algorithms, such as \cite{song2016deep,schroff2015facenet,Chen_2017_CVPR,bai2017regularized,bai2017ensemble}, is the design of a loss function that can learn more discriminant features. Since neural networks are usually trained using the stochastic gradient descent (SGD) in mini-batches, these loss functions are difficult to approximate the target of metric learning - pull samples with the same label into nearby points and push samples with different labels apart. \\

\noindent\textbf{Informative Sample Selection.} Given $N$ training images, there are about $O(N^2)$ pairs, $O(N^3)$ triplets, and $0(N^4)$ quadruplets. It is infeasible to traverses all these training tuples during training. Schroff {\em at. el.} \cite{schroff2015facenet} constructed a mini-batch of with 45 identities and each of which has 40 images. There are totally 1800 images in a mini-batch, and the approach obtained the state-of-art results on LFW face recognition challenge \cite{LFWTech}. While it is rather inconvenient to take thousands of images in a mini-batch with a large-scale network, due to the limitation of GPU memory. For deep metric learning, it is of great importance to selecting informative training tuples. Hard negative mining \cite{bucher2016hard} is widely used to select hard training tuples. Our work is closely related to that of  \cite{Wu_2017_ICCV,Harwood_2017_ICCV} which inspired the current work. Distance distribution was applied to guide tuple sampling for deep metric learning \cite{Wu_2017_ICCV,Harwood_2017_ICCV}. In this work, we strive to a further step by constructing a hierarchical tree that aggregates class-level global context, and formulating tuple selection elegantly by introducing a new violate margin. \\


\begin{figure*}[t]
  \centering
  \includegraphics[height=3.8cm,width=11cm]{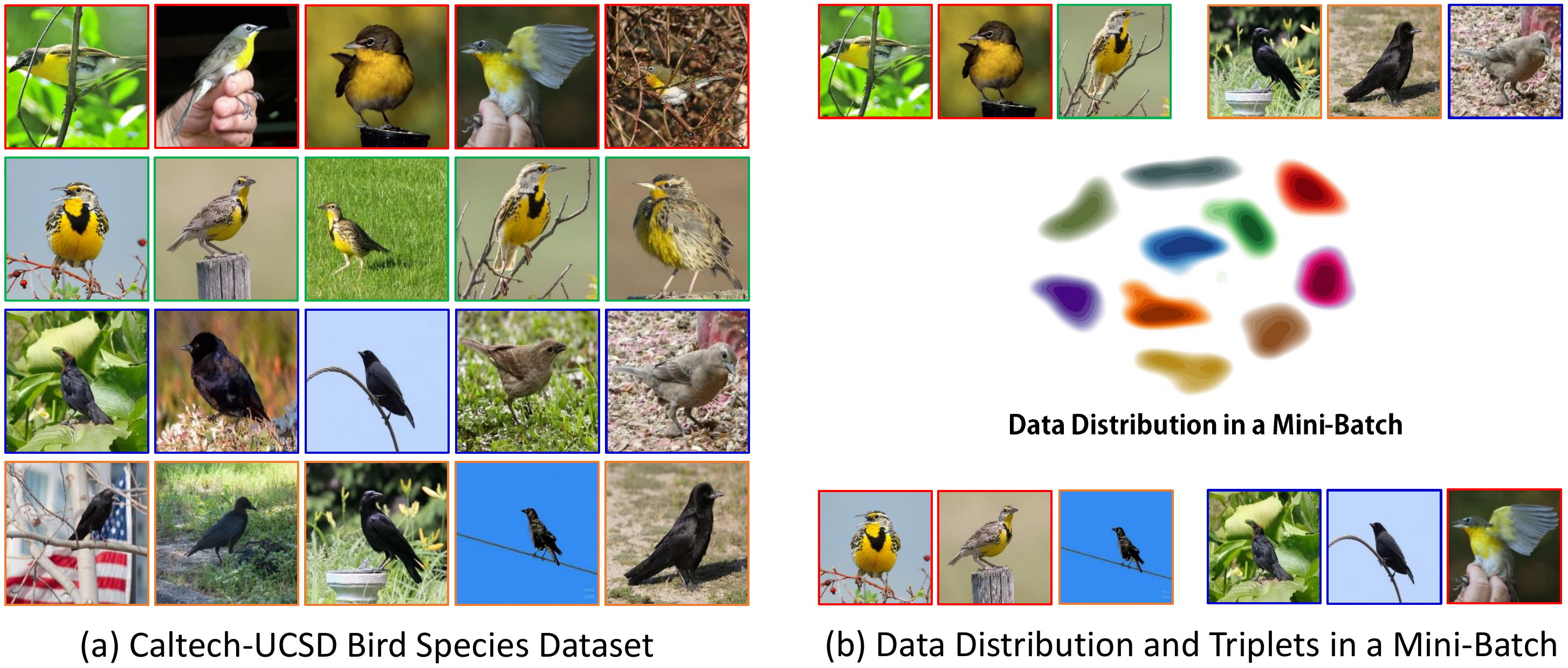}
  \caption{(a) Caltech-UCSD Bird Species Dataset \cite{wah2011caltech}. Images in each row are from the same class. There are four classes in different colors --- red, green, blue and yellow. (b) Data distribution and triplets in a mini-batch. Triplets in the top row violate the triplet constrain in the traditional triplet loss. Triplets in the bottom row are ignored in the triplet loss, but are revisited in the hierarchical triplet loss.}
  \label{Fig:Triplet loss and Hierarchical Loss}
\end{figure*}
    \section{Motivation: Challenges in Triplet Loss}
We start by revisiting the main challenges in standard triplet loss \cite{schroff2015facenet}, which we believe have a significant impact to the performance of deep triplet embedding. 

\subsection{Preliminaries}
Let $(\boldsymbol{x}_i, y_i)$ be the $i$-th sample in the training set $\mathcal{D} = \left \{  \left (  \boldsymbol{x}_i, y_i \right ) \right \}_{i=1}^{N}$. The feature embedding of $\boldsymbol{x}_i$ is represented as $\phi \left ( \boldsymbol{x}_i, \boldsymbol{\theta } \right ) \in \mathbb{R}^d$, where $\boldsymbol{\theta }$ is the learnable parameters of a differentiable deep networks, $d$ is the dimension of embedding and $y_i$ is the label of $\boldsymbol{x}_i$. $\phi \left ( \cdot, \boldsymbol{\theta } \right )$ is usually normalized into unit length for the training stability and comparison simplicity as in \cite{schroff2015facenet}. During the neural network training, training samples are selected and formed into triplets, each of which $\mathcal{T}_z = \left ( \boldsymbol{x}_a,\boldsymbol{x}_p,\boldsymbol{x}_n \right )$ are consisted of an anchor sample $\boldsymbol{x}_a$, a positive sample $\boldsymbol{x}_p$ and a negative sample $\boldsymbol{x}_n$. The labels of the triplet $\mathcal{T}_z = \left ( \boldsymbol{x}_a^z,\boldsymbol{x}_p^z,\boldsymbol{x}_n^z \right )$ satisfy $y_a = y_p \neq y_n$. Triplet loss aims to pull samples belonging to the same class into nearby points on a manifold surface, and push samples with different labels apart from each other. The optimization target of the triplet $\mathcal{T}_z$ is,
\begin{equation*}
    \begin{aligned}
        l_{tri}\left ( \mathcal{T}_z \right ) = \frac{1}{2} \left [ \left \| \boldsymbol{x}_a^z - \boldsymbol{x}_p^z  \right \| ^ 2 -  \left \| \boldsymbol{x}_a^z - \boldsymbol{x}_n^z  \right \| ^ 2 + \alpha \right ]_{+}.
    \end{aligned}
    \label{eq:triplet loss}
\end{equation*}

$\left [ \cdot \right ]_{+} =$ max$\left ( 0,\cdot \right )$ denotes the hinge loss function, and $\alpha$ is the violate margin that requires the distance $\left \| \boldsymbol{x}_a^z - \boldsymbol{x}_n^z  \right \| ^ 2$ of negative pairs to be larger than the distance $\left \| \boldsymbol{x}_a^z - \boldsymbol{x}_p^z  \right \| ^ 2$ of positive pairs. For all the triplets $\mathcal{T}$ in the training set $\mathcal{D} = \left \{  \left (  \boldsymbol{x}_i, y_i \right ) \right \}_{i=1}^{N}$, the final objective function to optimize is,

    \begin{equation*}
    \begin{aligned}
        \mathcal{L} = \frac{1}{Z} \sum _{\mathcal{T}^z \in \mathcal{T}} l_{tri}\left ( \mathcal{T}_z \right ),
    \end{aligned}
    \label{eq:triplet objective function}
    \end{equation*}
where $Z$ is the normalization term. For training a triplet loss in deep metric learning, the violate margin plays a key role to sample selection.

\subsection{Challenges}
\noindent\textbf{Challenge 1: triplet loss with random sampling.}
For many deep metric learning loss functions, such as contrastive loss \cite{hadsell2006dimensionality}, triplet loss \cite{schroff2015facenet} and  quadruplet loss \cite{Chen_2017_CVPR}, all training samples are treated equally with a constant violate margin, which only allows training samples that violate this margin to produce gradients. For a training set $\mathcal{D} = \left \{  \left (  \boldsymbol{x}_i, y_i \right ) \right \}_{k=1}^{N}$ with $N$ samples, training a triplet loss will generate $O\left ( N^3 \right )$ triplets, which is infeasible to put all triplets into a single mini-batch. When we sample the triplets over the whole training set randomly, it has a risk of slow convergence and pool local optima.
 We identify the problem that most of training samples obey the violate margin when the model starts to converge. These samples can not contribute gradients to the learning process, and thus are less informative, but can dominate the training process, which significantly degrades the model capability, with a slow convergence. This inspired current work that formulates the problem of sample selection via setting a dynamic violate margin, which allows the model to focus on a small set of informative samples.

 However, identifying informative samples from a vast number of the generated triplets is still challenging. This inspires us to strive to a further step, by sampling meaningful triplets from a structural class tree, which defines class-level relations over all categories. This transforms the problem of pushing hard samples apart from each other into encouraging a larger distance between two confusing classes. This not only reduces the search space, but also avoid over-fitting the model over individual samples, leading to a more discriminative model that generalizes better.  \\


\noindent\textbf{Challenge 2: risk of local optima.} Most of the popular metric learning algorithms, such as the contrastive loss, the triplet loss, and the quadruplet loss, describe similarity relationship between individual samples locally in a mini-batch, without considering global data distribution. In triplet loss, all triplet is treated equally. As shown in Fig. \ref{Fig:Triplet loss and Hierarchical Loss}, when the training goes after several epoches, most of training triplets dose not contribute to the gradients of learnable parameters in deep neural networks. There has been recent work that aims to solve this problem by re-weighting the training samples, as in \cite{wu2017sampling}. However, even with hard negative mining or re-weighting, the triplets can only see a few samples within a mini-batch, but not the whole data distribution. It is difficult for the triplet loss to incorporate the global data distribution on the target manifold space.
Although the data structure in the deep feature space are changed dynamically during the training process, the relative position of data points can be roughly preserved. This allows us to explore the data distribution obtained in the previous iterations to guide sample selection in the current stage. With this prior knowledge of data structure, a triplet, which does not violate the original margin $\alpha$,  is possible to generate gradients that contribute to the network training, as shown in Fig. \ref{Fig:Triplet loss and Hierarchical Loss}. Discriminative capability can be enhanced by learning from these hard but informative triplets.



	\section{Hierarchical Triplet Loss}

We describe details of the proposed hierarchical triplet loss, which contains two main components, constructing a hierarchical class tree and formulating the hierarchical triplet loss with a new violate margin. The hierarchical class tree is designed to capture global data context, which is encoded into triplet sampling via the new violate margin, by formulating the hierarchical triplet loss.

\begin{figure*}[ht]
  \centering
  \includegraphics[height=4cm,width=11cm]{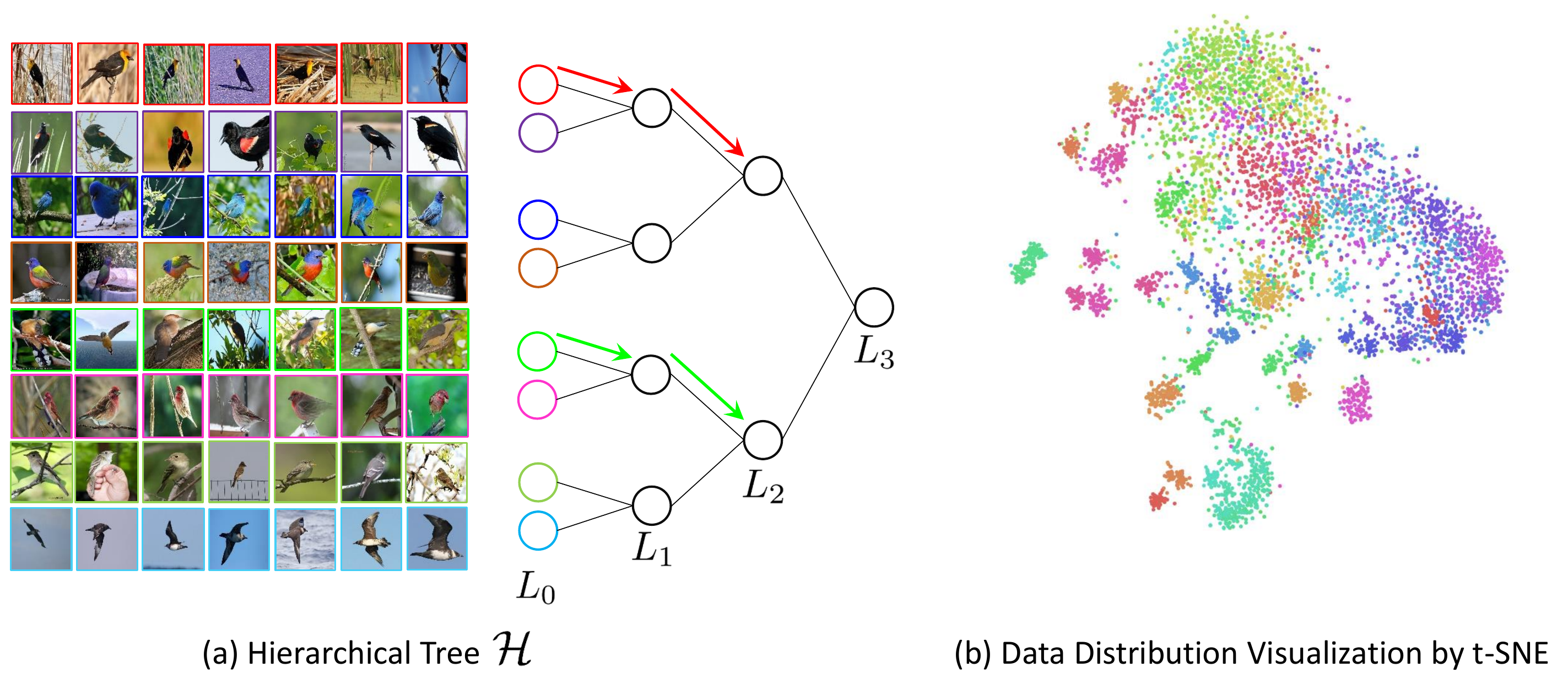}
  \caption{(a) A toy example of the hierarchical tree $\mathcal{H}$. Different colors represent different image classes in CUB-200-2011 \cite{wah2011caltech}. The leaves are the image classes in the training set. Then they are merged recursively until to the root node. (b) The training data distribution of 100 classes visualized by using t-SNE \cite{maaten2008visualizing} to reduce the dimension of triplet embedding from 512 to 2.}
  \label{Fig:Hierarchical Tree}
\end{figure*}

\subsection{Manifold Structure in Hierarchy}
We construct a global hierarchy at the class level. Given a neural network $\phi _t \left ( \cdot, \boldsymbol{\theta} \right ) (\in {\mathbb{R}}^d)$ pre-trained using the traditional triplet loss, we get the hierarchical data structure based on sample rules. Denote the deep feature of a sample $\boldsymbol{x}_i$ as $\boldsymbol{r}_i = \phi _t \left ( \boldsymbol{x}_i, \boldsymbol{\theta} \right )$. We first calculate a distance matrix of $\mathcal{C}$ classes in the whole training set $\mathcal{D}$. The distance between the $p$-th class and the $q$-th class is computed as,

    \begin{equation*}
    \begin{aligned}
        d\left ( p, q \right ) = \frac{1}{n_p n_q} \sum _{i \in p, j \in q} \left \| \boldsymbol{r}_i - \boldsymbol{r}_j \right \| ^ 2,
    \end{aligned}
    \label{eq:triplet objective function}
    \end{equation*}
where $n_p$ and $n_q$ are the numbers of training samples in the $p$-th and the $q$-th classes respectively. Since the deep feature $\boldsymbol{r}_i$ is normalized into unit length, the value of the interclass distance $d\left ( p, q \right )$ varies from 0 to 4.

We build hierarchical manifold structure by creating a hierarchical tree, according to the computed interclass distances. The leaves of the hierarchical tree are the original image classes, where each class represents a leave node at the $0$-th level. Then hierarchy is created by recursively merging the leave notes at different levels, based on the computed distance matrix. The hierarchical tree is set into $L$ levels, and the average inner distance $d_0$ is used as the threshold for merging the nodes at the $0$-th level.

    \begin{equation*}
    \begin{aligned}
        d_0 = \frac{1}{\mathcal{C}} \sum _{c=1} ^{\mathcal{C}} \left ( \frac{1}{n_c ^2 -n_c} \sum _{i \in c, j \in c} \left \| \boldsymbol{r}_i - \boldsymbol{r}_j  \right \| ^2 \right ).
    \end{aligned}
    \label{eq:triplet objective function}
    \end{equation*}
where $n_c$ is the number of samples in the $c$-th class. Then the nodes are merged with different thresholds. At the $l$-th level of the hierarchical tree, the merging threshold is set to $d_l = \frac{l\left ( 4 - d_0 \right )}{L} + d_0$. Two classes with a distance less than $d_l$ are merged into a node at the $l$-th level. The node number at the $l$-th level is $N_l$. The nodes are merged from the $0$-th level to the $L$-th level. Finally, we generate a hierarchical tree $\mathcal{H}$ which starts from the leave nodes of original image classes to a final top node, as shown in Fig. \ref{Fig:Hierarchical Tree} (a). The constructed hierarchical tree captures class relationships over the whole dataset, and it is updated interactively at the certain iterations over the training.

\subsection{Hierarchical Triplet Loss}
We formulate the problem of triplet collection into a hierarchical triplet loss. We introduce a dynamical violate margin, which is the main difference from the conventional triplet loss using a constant violate margin. \\

\noindent\textbf{Anchor neighbor sampling.} We randomly select $l^{\prime}$ nodes at the $0$-th level of the constructed hierarchical tree $\mathcal{H}$. Each node represents an original class, and collecting classes at the $0$-th level aims to  preserve the diversity of training samples in a mini-batch, which is important for training deep networks with batch normalization \cite{ioffe2015batch}. Then $m - 1$ nearest classes at the $0$-th level are selected for each of the $l^{\prime}$ nodes, based on the distance between classes computed in the feature space. The goal of collecting nearest classes is to encourage model to learn discriminative features from the visual similar classes. Finally, $t$ images for each class are randomly collected, resulting in $n~(n= l^{\prime} m t)$ images in a mini-batch $\mathcal{M}$. Training triplets within each mini-batch  are generated from the collected $n$ images based on class relationships. We write the anchor-neighbor sampling into A-N sampling for convenience. \\

\begin{figure*}[ht]
  \centering
  \includegraphics[height=4cm, width=11cm]{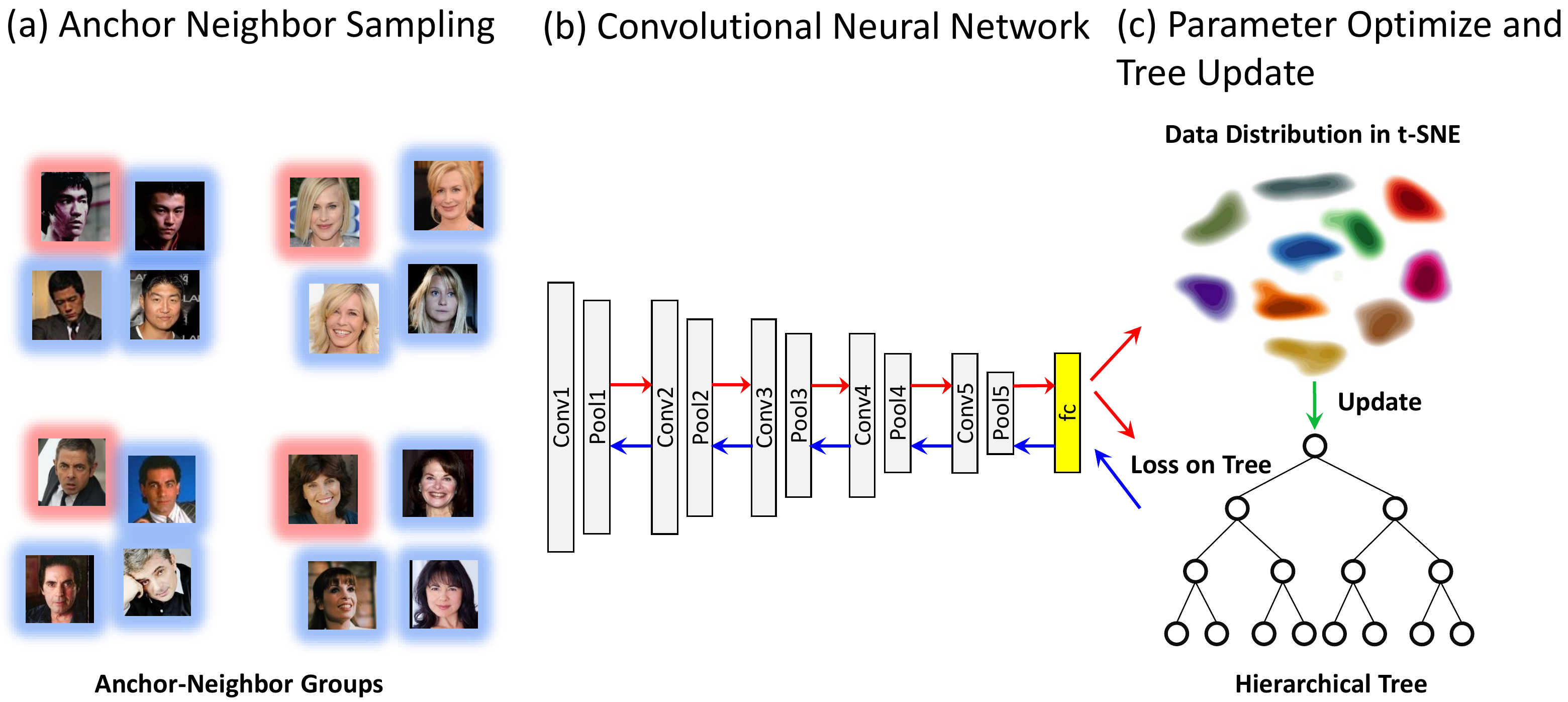}
  \caption{(a)Sampling strategy of each mini-batch. The images in red stand for anchors and the images in blue stand for the nearest neighbors. (b) Train CNNs with the hierarchical triplet loss. (c) Online update of the hierarchical tree.}
  \label{Fig:Overview}
\end{figure*}

\noindent\textbf{Triplet generation and dynamic violate margin.}
Hierarchical triplet loss (computed on a mini-batch of $\mathcal{M}$ ) can be formulated as,

    \begin{equation*}
    \begin{aligned}
        \mathcal{L}_{\mathcal{M}} = \frac{1}{2 Z_{\mathcal{M}}} \sum _{\mathcal{T}^z \in \mathcal{T}^{\mathcal{M}}} \left [ \left \| \boldsymbol{x}_a^z - \boldsymbol{x}_p^z  \right \| -  \left \| \boldsymbol{x}_a^z - \boldsymbol{x}_n^z  \right \| + \alpha _z \right ]_{+}.
    \end{aligned}
    \label{eq:adaptive triplet objective function}
    \end{equation*}
where $\mathcal{T}^{\mathcal{M}}$ is all the triplets in the mini-batch $\mathcal{M}$, and $Z_{\mathcal{M}} = A_{l^{\prime} m} ^ 2 A_t ^ 2 C_t^1$ is the number of triplets. Each triplet is constructed as $\mathcal{T}_z = \left ( \boldsymbol{x}_a,\boldsymbol{x}_p,\boldsymbol{x}_n \right )$, and the training triplets are generated as follows. $A_{l^{\prime} m} ^ 2$ indicates randomly selecting two classes - a positive class and a negative class, from all $l^{\prime} m$ classes in the mini-batch. $A_t ^ 2$ means selecting two samples - a anchor sample ($\boldsymbol{x}_a^z$) and a positive sample ($\boldsymbol{x}_p^z$), from the positive class, and $C_t ^ 1$ means randomly selecting a negative sample ($\boldsymbol{x}_n^z$) from the negative class. $A_{l^{\prime} m} ^ 2$, $A_t ^ 2$ and $C_t ^ 1$ are notations in combinatorial mathematics. See reference~\cite{van2001course} for details. \\

 $\alpha _{z}$ is a dynamic violate margin, which is different from the constant margin of traditional triplet loss. It is computed according to the class relationship between the anchor class $y_a$ and the negative class $y_n$ over the constructed hieratical class tree. Specifically, for a triplet $\mathcal{T}_z$, the violate margin $\alpha _{z}$ is computed as,
    \begin{equation*}
    \begin{aligned}
        \alpha _{z} = \beta  + d_{\mathcal{H}\left ( y_a, y_n \right )} -  s _{y_a},
    \end{aligned}
    \label{eq:triplet objective function}
    \end{equation*}
where $\beta~(= 0.1)$ is a constant parameter that encourages the image classes to reside further apart from each other than the previous iterations. $\mathcal{H}\left ( y_a, y_n \right )$ is the hierarchical level on the class tree, where the class $y_a$ and the class $y_n$ are merged into a single node in the next level. $d_{\mathcal{H}\left ( y_a, y_n \right )}$ is the threshold for merging the two classes on $\mathcal{H}$, and $s _{y_a} = \frac{1}{n_{y_a} ^2 -n_{y_a}} \sum _{i,j \in y_a} \left \| \boldsymbol{r}_i - \boldsymbol{r}_j  \right \| ^2$ is the average distance between samples in the class $y_a$. In our hierarchical triplet loss, a sample $\boldsymbol{x}_a$ is encouraged to push the nearby points with different semantic meanings apart from itself. Furthermore, it also contributes to the gradients of data points which are very far from it, by computing a dynamic violate margin which encodes global class structure via $\mathcal{H}$. For every individual triplet, we search on $\mathcal{H}$ to encode the context information of the data distribution for the optimization objective. Details of training process with the proposed hierarchical triplet loss are described in Algorithm 1.\\

\noindent\textbf{Implementation Details.}
All our experiments are implemented using Caffe~\cite{jia2014caffe} and run on an NVIDIA TITAN X(Maxwell) GPU with 12GB memory. The network architecture is a GoogLeNet \cite{szegedy2015going} with batch normalization \cite{ioffe2015batch} which is pre-trained on the ImageNet dataset \cite{ILSVRC15}. The 1000-way fully connected layer is removed, and replace by a $d$ dimensional fully connected layer. The new added layer is initialized with random noise using the "Xaiver" filler.  We modify the memory management of Caffe~\cite{jia2014caffe} to ensure it can take 650 images in a mini-batch for GoogLeNet with batch normalization. The input images are resized and cropped into $224 \times 224$, and then subtract the mean value. The optimization method used is the standard SGD with a learning rate $1{e^{-3}}$.

\begin{algorithm}[H]
\caption{Training with hierarchical triplet loss}
\LinesNumbered 
\KwIn{Training data $\mathcal{D} = \left \{  \left (  \boldsymbol{x}_i, y_i \right ) \right \}_{k=1}^{N}$. Network $\phi \left ( \cdot, \boldsymbol{\theta } \right )$ is initialized with a pretrained ImageNet model. The hierarchical class tree $\mathcal{H}$ is built according to the features of the initialized model. The margin $\alpha_z$ for any pair of classes is set to 0.2 at the beginning.}
\KwOut{The learnable parameters $\theta$ of the neural network $\phi \left ( \cdot, \boldsymbol{\theta } \right )$.}

\While{not converge}{
　　$t \leftarrow t+1$ \;
   Sample anchors randomly and their neighborhoods according to $\mathcal{H}$ \;
    Compute the violate margin for different pairs of image classes  by searching through the hierarchical tree $\mathcal{H}$ \;
   Compute the hierarchical triplet loss in a mini-batch  $\mathcal{L}_{\mathcal{M}}$\;
   Backpropagate the gradients produced at the loss layer and update the learnable parameters \;
   At each epoch, update the hierarchical tree $\mathcal{H}$ with current model.
}
\end{algorithm}

%


\section{Experimental Results and Comparisons}

We evaluate the proposed hierarchical triplet loss on the tasks of image retrieval and face recognition. Extensive experiments are conducted on a number of benchmarks, including \emph{In-Shop Clothes Retrieval}~\cite{liu2016deepfashion} and \emph{Caltech-UCSD Birds 200}~\cite{wah2011caltech} for image retrieval, and LFW~\cite{LFWTech} for face verification. Descriptions of dataset and implementation details are presented as follows.

\begin{table*}[t]\small
\setlength{\abovecaptionskip}{10pt}
\setlength{\belowcaptionskip}{-10pt}
\begin{center}
\resizebox{0.65\textwidth}{!}
{
\begin{tabular}{@{}lccccccc@{}}
\toprule
R@                                    & 1         &10      &20       &30      &40        &50     \\ \midrule
FashionNet+Joints\cite{liu2016deepfashion}  & 41.0      &64.0    &68.0     &71.0    &73.0      &73.5   \\
FashionNet+Poselets\cite{liu2016deepfashion}& 42.0      &65.0    &70.0     &72.0    &72.0      &75.0   \\
FashionNet\cite{liu2016deepfashion}         & 53.0      &73.0    &76.0     &77.0    &79.0      &80.0   \\
HDC\cite{Yuan_2017_ICCV}                & 62.1      &84.9    &89.0     &91.2    &92.3      &93.1   \\
BIER\cite{opitz2017bier}               & 76.9      &92.8    &95.2     &96.2    &96.7     &97.1   \\\midrule
Ours Baseline                         & 62.3      &85.1  &89.0          &91.1  &92.4    &93.4   \\
A-N Sampling              & 75.3        &91.8  &94.3          &96.2  &96.7           &97.5\\
HTL            &\textbf{80.9}  &\textbf{94.3}  &\textbf{95.8}          &\textbf{97.2}          &\textbf{97.4} &\textbf{97.8}\\
\bottomrule
\end{tabular}
}
\end{center}
\caption{Comparisons on the In-Shop Clothes Retrieval dataset ~\cite{liu2016deepfashion}.}
\label{In-Shop Clothes dataset}
\end{table*}

\begin{figure*}[ht]
  \centering
  \includegraphics[height=6cm,width=12cm]{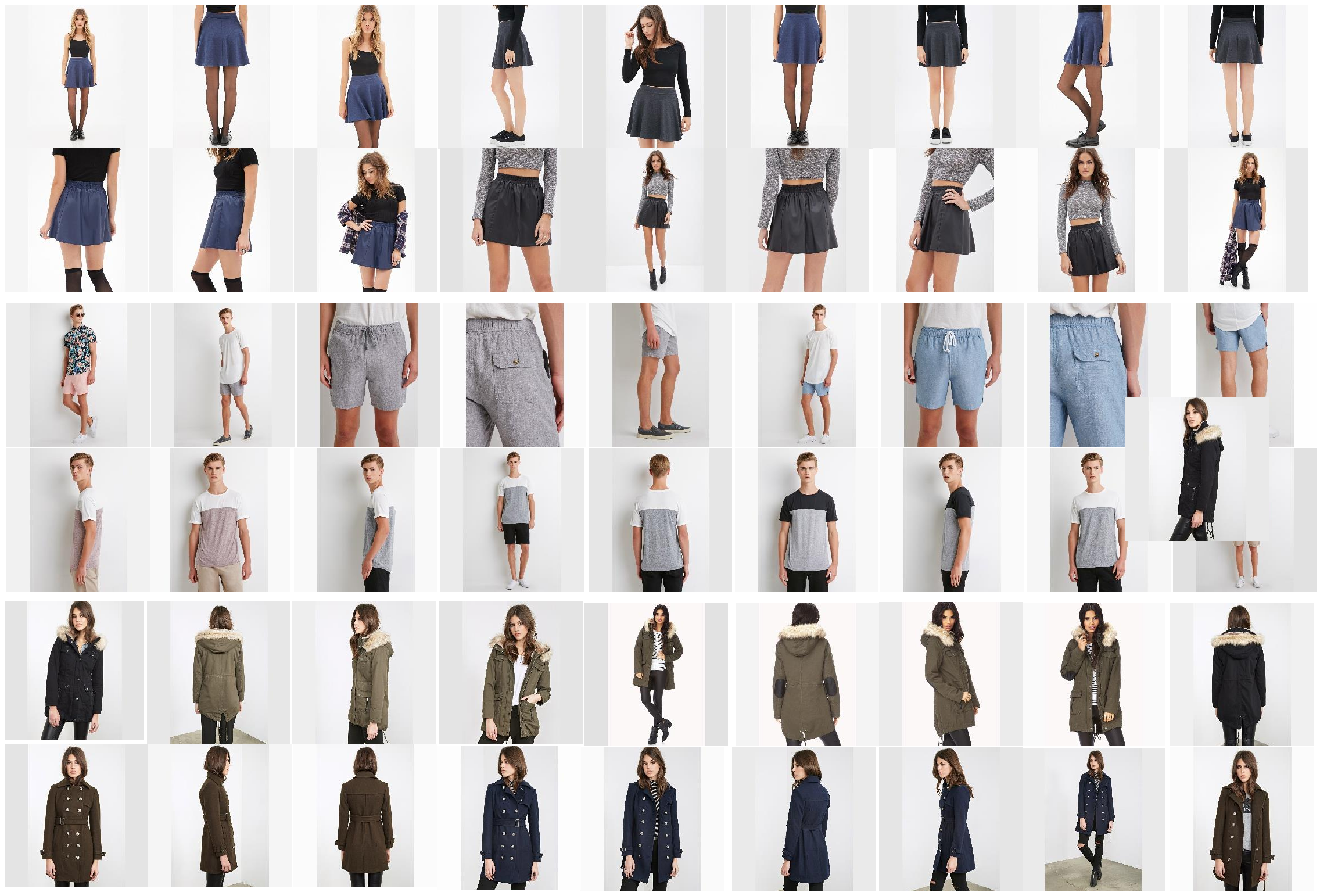}
  \caption{Anchor-Neighbor visualization on \emph{In-Shop Clothes Retrieval} training set ~\cite{liu2016deepfashion}. Each row stands for a kind of fashion style. The row below each odd row is one of neighborhoods of the fashion style in the odd row.}
  \label{Fig:Anchor Neighborhood Sampling}
\end{figure*}

\subsection{In-Shop Clothes Retrieval}

\noindent\textbf{Datasets and performance measures.} The \emph{In-Shop Clothes Retrieval} dataset ~\cite{liu2016deepfashion} is very popular in image retrieval. It has 11735 classes of clothing items and 54642 training images. Following the protocol in \cite{liu2016deepfashion,Yuan_2017_ICCV}, 3997 classes are used for training (25882 images) and 3985 classes are for testing (28760 images). The test set are partitioned into the query set and the gallery set, both of which has 3985 classes. The query set has 14218 images and the gallery set has 12612 images. As in Fig. \ref{Fig:Anchor Neighborhood Sampling}, there are a lot image classes that have very similar contents.

For the evaluation, we use the most common Recall$@K$ metric. We extract the features of each query image and search the $K$ most similar images in the gallery set. If one of the $K$ retrieved images have the same label with the query image, the recall will increase by 1, otherwise will be 0. We evaluate the recall metrics with $K \in \left \{ 1, 2, 4, 8, 16, 32 \right \}$.

\noindent\textbf{Implementation details.} Our network is based on GoogLeNet V2 \cite{ioffe2015batch}. The dimension $d$ of the feature embedding is 128. The triplet violate margin is set to 0.2. The hierarchical tree has 16 levels including the leaves level which contains the images classes. At the first epoch, the neural network is trained with the standard triplet loss which samples image classes for mini-batches randomly. Then during the training going on, the hierarchical tree is updated and used in the following steps. Since there are 3997 image classes for training and there many similar classes, the whole training needs 30 epoch and the batch size is set to 480. For every 10 epoch, we decrease the learning rate by multiplying 0.1. The testing codes are gotten from HDC \cite{Yuan_2017_ICCV}.

\noindent\textbf{Result comparison.} We compare our method with existing state-of-the-art algorithms and our baseline --- triplet loss. Table~\ref{In-Shop Clothes dataset} lists the results of image retrieval on \emph{In-Shop Clothes Retrieval}. The proposed method achieves 80.9\% Recall$@1$, and outperforms the baseline algorithm --- triplet loss by 18.6\%. It indicates that our algorithm can improve the discriminative power of the original triplet loss by a large margin. State-of-the-art algorithms, including HDC~\cite{Yuan_2017_ICCV}, and BIER~\cite{opitz2017bier}, used boosting and ensemble method to take the advantage of different features and get excellent results. Our method demonstrates that by incorporate the global data distribution into deep metric learning, the performance will be highly improved. The proposed hierarchical loss get 80.9\% Recall$@1$, which is 4.0\% higher than BIER~\cite{opitz2017bier} and 18.8\% higher than HDC~\cite{Yuan_2017_ICCV}.

\subsection{Caltech-UCSD Birds 200-2011}
\begin{table*}[t]\small
\setlength{\abovecaptionskip}{10pt}
\setlength{\belowcaptionskip}{-10pt}
\begin{center}
\resizebox{0.6\textwidth}{!}
{
\begin{tabular}{@{}lccccccc@{}}
\toprule
R@                                       & 1         &2      &4       &8      &16        &32     \\ \midrule
LiftedStruct\cite{song2016deep}                 & 47.2      &58.9   &70.2    &80.2   &89.3      &93.2   \\
Binomial Deviance\cite{ustinova2016learning}                 & 52.8      &64.4   &74.7    &83.9   &90.4      &94.3   \\
Histogram Loss\cite{ustinova2016learning}                 & 50.3      &61.9   &72.6    &82.4   &88.8      &93.7   \\
N-Pair-Loss\cite{sohn2016improved}                 & 51.0      &63.3   &74.3    &83.2   &-         &-   \\
HDC\cite{Yuan_2017_ICCV}                 & 53.6      &65.7   &77.0    &85.6   &91.5      &95.5   \\
BIER\cite{opitz2017bier}                 & 55.3      &67.2   &76.9    &85.1   &91.7      &95.5   \\\midrule
Ours Baseline                            & 55.9      &68.4   &78.2    &86.0   &92.2      &95.5   \\
HTL                                &\textbf{57.1}      &\textbf{68.8}   &\textbf{78.7}    &\textbf{86.5}   &\textbf{92.5}      &\textbf{95.5}   \\
\bottomrule
\end{tabular}
}
\end{center}
\caption{Comparison with the state-of-art on the CUB-200-2011 dataset \cite{wah2011caltech}.}
\label{cub dataset}
\end{table*}
\noindent\textbf{Datasets and performance measures.} The \emph{Caltech-UCSD Birds 200} dataset (CUB-200-2011) ~\cite{wah2011caltech} contains photos of 200 bird species with 11788 images.  CUB-200-2011 serves as a benchmark in most existing work on deep metric learning and image retrieval. The first 100 classes (5864 images) are used for training, and the rest (5924 images) of classes are used for testing. The rest images are treated as both the query set and the gallery set. For the evaluation, we use the same Recall$@K$ metric as in Section \emph{In-Shop Clothes Retrieval}. Here, $K \in \left \{ 1, 2, 4, 8, 16, 32 \right \}$.

\noindent\textbf{Implementation details.} The dimension $d$ of the feature embedding is 512. The triplet violate margin is set to 0.2. As in the previous section, the hierarchical tree is still set to 16 levels. All the training details are almost the same with the \emph{In-Shop Clothes Retrieval} dataset. But since there are only 100 image classes for training, the dataset is very easy to get overfitting. When we train 10 epoches, the training stopped. The batch size is set to 50. For every 3 epoch, we decrease the learning rate by multiplying 0.1.

\noindent\textbf{Result comparison.} Table~\ref{cub dataset} lists the results of image retrieval on \emph{Caltech-UCSD Birds 200-2011}. The baseline --- triplet loss already get the state-of-art results with 55.9\% Recall$@1$ compared with the previous state-of-art HDC 54.6\% and BIER 55.3\%. If we use the anchor-neighbor sampling and the hierarchical loss, we get 57.1\% Recall$@1$. Since there are only 100 classes and 6000 images for training, the network is very easy to get overfitting. The performance gain gotten by the hierarchical loss is only 1.2\% Recall$@1$.

\subsection{Cars-196~\cite{Krause2013} and Stanford Online Products~\cite{song2016deep}}\vspace{-2mm}
Details of the Cars-196 and Stanford Online Products~\cite{song2016deep} are described in ~\cite{Krause2013,song2016deep}. The dimension of the feature embedding is set to 512. The triplet violate margin is set to 0.2, with a hierarchical tree of $depth=16$. The whole training needs 30 epoch and the batch size is set to 50. For every 10 epoch, we decrease the learning rate by multiplying 0.1.

Results are presented in Table \ref{my-label2}, where the proposed HTL outperforms our baseline, BIER and HDC, with clear margins on both datasets. Specifically, on the Cars-196, HTL achieves 81.4\% Recall$@1$, which outperforms orginal triplet loss by 2.2\%, and previous state-of-art by 3.4\%. On the Stanford Online Products, HTL achieves 74.8\% Recall$@1$, outperforming triplet loss by 2.2\%, and previous state-of-art by 2.1\%. These results demonstrate that the proposed HTL can improve original triplet loss efficiently, and further proved the generalization ability of HTL.
\vspace{-2em}
\begin{table*}[]\small
\setlength{\abovecaptionskip}{-5pt}
\setlength{\belowcaptionskip}{-9pt}
\centering
\begin{center}
\resizebox{0.85\textwidth}{!}
{
\begin{tabular}{l|cccccc|cccc}
\hline
\hline
              & \multicolumn{6}{c|}{Cars-196}           & \multicolumn{4}{c}{Stanford Online Products} \\ \hline
R@            & 1    & 2    & 4    & 8    & 16   & 32   & 1         & 10        & 100       & 100       \\ \hline
HDC           & 73.7 & 83.2 & 89.5 & 93.8 & 96.7 & 98.4 & 69.5      & 84.4      & 92.8      & 97.7      \\ \hline
BIER          & 78.0 & 85.8 & 91.1 & 95.1 & 97.3 & 98.7 & 72.7      & 86.5      & 94.0      & 98.0      \\ \hline
Baseline      & 79.2 & 87.2 & 92.1 & 95.2 & 97.3 & 98.6 & 72.6      & 86.2      & 93.8      & 98.0      \\ \hline
HTL(depth=16) & \textbf{81.4} & \textbf{88.0} & \textbf{92.7} & \textbf{95.7} & \textbf{97.4} & \textbf{99.0} & \textbf{74.8}      & \textbf{88.3}      & \textbf{94.8}      & \textbf{98.4}      \\ \hline
\hline
\end{tabular}
}
\end{center}
\caption{Comparison with the state-of-art on the cars-196 and Stanford products.}
\label{my-label2}
\end{table*}

\subsection{LFW Face Verification}
\noindent\textbf{Datasets and performance measures.} The \emph{CASIA-WebFace} dataset ~\cite{yi2014learning} is one of the publicly accessible datasets for face recognition. It has been the most popular dataset for the training of face recognition algorithms, such as in \cite{amos2016openface,wen2016discriminative,liu2017sphereface}. \emph{CASIA-WebFace} has 10575 identities and 494414 images. We following the testing protocol in ~\cite{yi2014learning} to test the performance of our algorithms. The face verification results on LFW dataset \cite{LFWTech} is reported.

\noindent\textbf{Implementation details.} Since the triplet loss is very sensitive to the noise, we clear the CASIA-WebFace using the pre-trained model of VGG-Face~\cite{parkhi2015deep} and manually remove some noises. About $10\%$ images are removed. Then the remained faces are used to train a SoftMax classifier. The network parameters are initialized by a pre-trained ImageNet model. We fine-tune the pre-trained classification network for face recognition using the hierarchical loss.

\noindent\textbf{Result comparison.} The triplet loss gets $98.3\%$ accuracy on the LFW face verification task, which is $1.12\%$ lower than the SpereFace\cite{liu2017sphereface} --- $99.42\%$ which uses the same dataset for training. When we substitute the triplet loss with the hierarchical triplet loss, the results comes to $99.2$. It's comparable with state-of-art results. This indicates that the hierarchical triplet loss has stronger discriminative power than triplet loss. While, since the triplet based method are very sensitive to noise, the hierarchical triplet loss get inferior performance compared with SphereFace~\cite{liu2017sphereface} $99.42\%$ and FaceNet~\cite{schroff2015facenet} $99.65\%$ .

\subsection{Sampling Matter and Local Optima}\vspace{-2mm}

\begin{figure*}[ht]
  \centering
  \includegraphics[height=4cm, width=12cm]{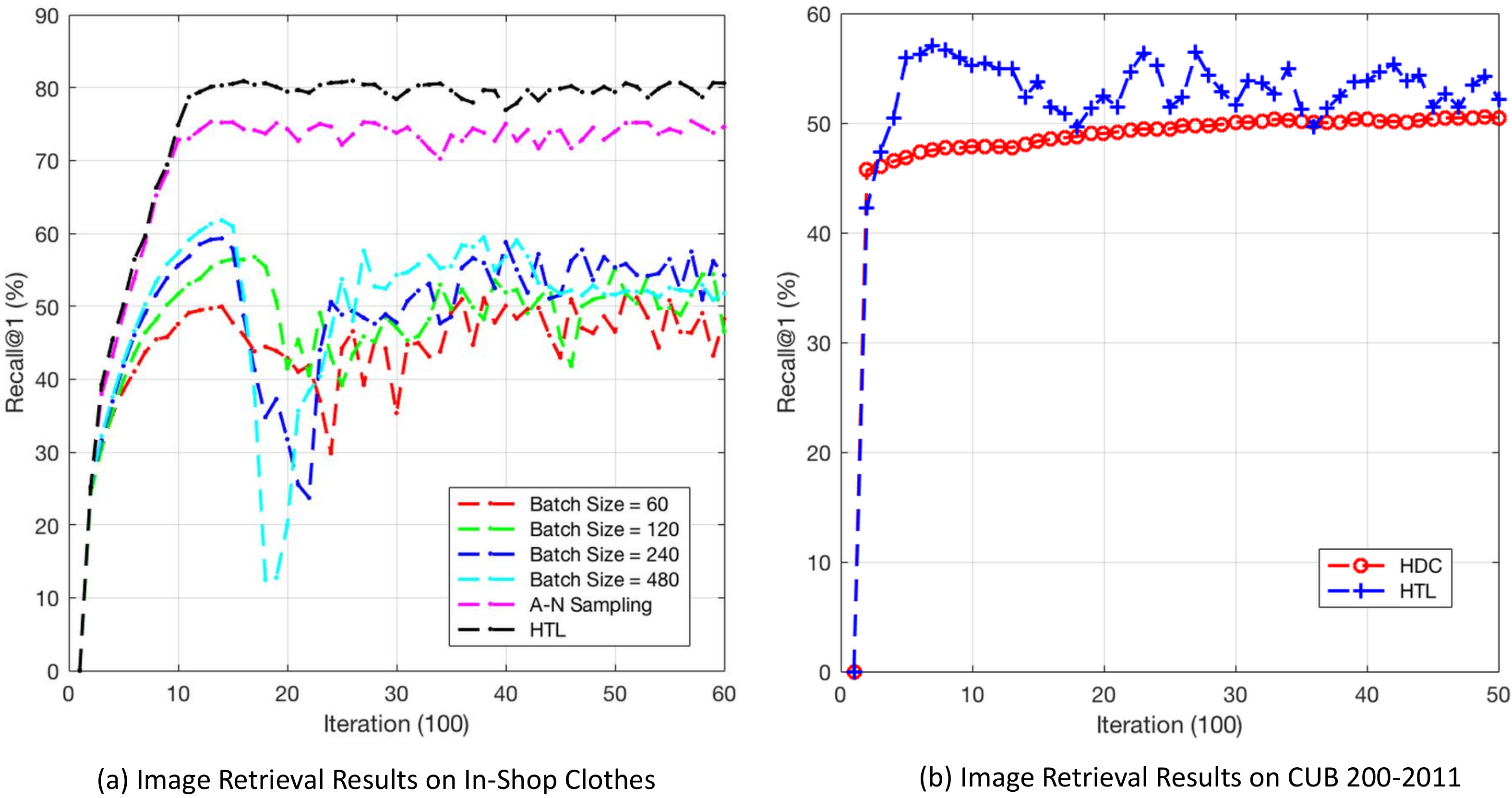}
  \caption{(a) Image retrieval results on In-Shop Clothes~\cite{liu2016deepfashion} with various batch sizes. (b) Image retrieval results on CUB-200-2011~\cite{wah2011caltech}.}
  \label{Fig:Convergence}
\end{figure*}

\noindent\textbf{Sampling Matter.} We investigate the influence of batch size on the test set of \emph{In-Shop Clothes Retrieval}. Fig. \ref{Fig:Convergence} (a) shows that when the batch size grows from 60 to 480, the accuracy increases in the same iterations. When the training continues, the performance will fluctuates heavily and get overfitting. Besides, when come to the same results at $60\%$ Recall@1, both the anchor-neighbor sampling with triplet loss and the hierarchical loss converge at about 2 times faster than random sampling (Batch Size = 480). Fig. \ref{Fig:Convergence} (b) shows the compares the convergence speed of the triplet loss (our baseline), the hierarchical triplet loss and the HDC~\cite{Yuan_2017_ICCV} on the test set of \emph{Caltech-UCSD Birds 200}. Compared to the 60000 iterations (see in \cite{Yuan_2017_ICCV}), the hierarchical triplet loss converges in 1000 iterations. The hierarchical triplet loss with anchor-Neighborhood sampling converge faster traditional and get better performance than HDC~\cite{Yuan_2017_ICCV}.

\noindent\textbf{Pool Local Optima.} In Table \ref{In-Shop Clothes dataset} and Table \ref{cub dataset}, we can find that the triplet loss get inferior performance than the hierarchical triplet loss on both the \emph{In-Shop Clothes Retrieval} and  \emph{Caltech-UCSD Birds 200}. In the Fig. \ref{Fig:Convergence}, the accuracy of the triplet loss start to fluctuate when the training continues going after the loss drops to very low. In fact, there are always very few or zeros triplets in mini-batch even when the network isn't gotten the best results. Then they don't produce gradients and will decay the learnable parameters in networks by SGD~\cite{orr2003neural}. So we incorporate the hierarchical structure to make points in the mini-batch know the position of point that are already far away, and then attempt to push them further from itself and its neighborhood classes. 


\subsection{Ablation Study}


We perform ablation studies on In-Shop Clothes and CUB-200-2011, as reported in Table \ref{ablation_study}. First, directly applying hard negative sampling (HNS) to the whole training set is difficult to obtain a performance gain. Actually, our baseline model applies a semi-HNS, which outperforms HNS.
We design a strong class-level constrain - Anchor-Neighbor Sampling of HTL, which encourages the model to learn discriminative features from visual similar classes. This is the key to performance boost. Second, we integrated the proposed anchor-neighbor sampling and dynamic violate margin into HDC  where a contrastive loss is used. As shown in Table \ref{ablation_study} (bottom), HDC+ got an improvement of 7.3\% R@1 on the In-Shop Clothes Retrieval, suggesting that our methods work practically well with a contrastive loss and HDC. Third, HTL with a depth of 16 achieves best performance at R@1 of 80.9\%. This is used as default setting in all our experiments. We also include results of ``flat'' tree with depth=1. Results suggest that the ``flat'' tree with the proposed dynamic violate margin improves the R@1 from 75.3\% to 78.9\%, and hierarchy tree improves it further to 80.9\%. \\
~\vspace{-8mm}
\begin{table*}[]
\setlength{\abovecaptionskip}{5pt}
\setlength{\belowcaptionskip}{0pt}
\centering
\resizebox{1.0\textwidth}{!}
{
\begin{tabular}{l|cccccc|cccccc}
\hline
\hline
                          & \multicolumn{6}{c|}{In-Shop Clothes}    & \multicolumn{6}{c}{CUB-200-2011}        \\ \hline
R@                        & 1    & 10   & 20   & 30   & 40   & 50   & 1    & 2    & 4    & 8    & 16   & 32   \\ \hline

\hline
\multicolumn{13}{c}{On Triplets with Sampling} \\\hline
Random Sampling& 59.3 & 83.5& 87.9 & 90.5 & 91.3 & 93.0 & 51.4 & 63.9 & 74.8 & 83.4 & 90.0 & 94.3 \\ \hline
Hard Negative Mining      & 60.1 & 84.3 & 88.2 & 90.2 & 91.5 & 92.6 & 51.6 & 63.9 & 74.2 & 84.4 & 89.9 & 94.6 \\ \hline
Semi-Hard Negative Mining                & 62.3 & 85.1 & 89.0 & 91.1 & 92.4 & 93.4 & 55.9 & 68.4 & 78.2 & 86.0 & 92.2 & 95.5 \\ \hline
Anchor-Neighbor Sampling (HTL)              & 75.3 & 91.8 & 94.3 & 96.2 & 96.7 & 97.5 & 56.4 & 68.5 & 78.5 & 86.2 & 92.4 & 95.5 \\ \hline
\hline

\multicolumn{13}{c}{HTL with A-N Sampling + Dynamic Violate Margin($\alpha_z$)} \\\hline
Class Proxy(flat/depth=1)         & 78.9 & 93.4 & 94.8 & 96.0 & 96.5 & 97.5 & 56.0 & 68.1 & 78.2 & 86.2 & 92.3 & 95.5 \\ \hline

HTL(depth=8)              & 78.7 & 93.3 & 94.6 & 96.2 & 96.9 & 97.4 & 56.2 & 68.5 & 78.3 & 86.1 & 92.3 & 95.5 \\ \hline
HTL(depth=16)             & \textbf{80.9} & \textbf{94.3} & \textbf{95.8} & \textbf{97.2} & \textbf{97.4} & \textbf{97.8} & \textbf{57.1} & \textbf{68.8} & \textbf{78.7} & \textbf{86.5} & \textbf{92.5} & \textbf{95.5} \\ \hline
HTL(depth=32)             & 79.3 & 93.8 & 95.0 & 96.9 & 97.1 & 97.5 & 56.4 & 68.5 & 78.5 & 86.2 & 92.3 & 95.5 \\ \hline
\hline

\multicolumn{13}{c}{HDC+: Contrastive Loss with A-N Sampling + Dynamic Violate Margin($\alpha_z$)} \\\hline
HDC                       & 62.1 & 84.9 & 89.0 & 91.2 & 92.3 & 93.1 & 53.6 & 65.7 & 77.0 & 85.6 & 91.5 &
95.5 \\ \hline
HDC+                      & 69.4 & 88.6 & 93.4 & 94.1 & 95.3 & 96.5 & 54.1 & 66.3 & 77.2 & 85.6 & 91.7 & 95.5 \\ \hline
 \hline
\end{tabular}
}
\caption{Ablation Studies on In-Shop Clothes Retrieval and CUB-200-2011.}
\label{ablation_study}
\end{table*}
\vspace{-2em}

	\section{Conclusion}\vspace{-3mm}
We have presented a new hierarchical triplet loss (HTL) which is able to
select informative training samples (triplets) via an adaptively-updated hierarchical tree that encodes global context. HTL effectively handles the main limitation of random sampling, which is a critical issue for deep metric learning. First, we construct a hierarchical
tree at the class level which encodes global context information over the whole dataset. Visual similar classes are merged recursively to form the hierarchy. Second, the problem of triplet collection is formulated by proposing a new violate margin, which is computed dynamically
based on the designed hierarchical tree. This allows it to learn from more meaningful hard samples with the guide of global context.
The proposed HTL is evaluated on the tasks of image retrieval and face recognition, where it achieves new state-of-the-art performance on a number of standard benchmarks.

\clearpage

\bibliographystyle{splncs04}
\bibliography{egbib}	
\end{document}